\documentclass[letterpaper, 10 pt, conference]{ieeeconf}  

\IEEEoverridecommandlockouts                              
\usepackage{graphicx}
\usepackage[utf8]{inputenc}
\usepackage[T1]{fontenc}
\usepackage[french]{babel}
\usepackage{amsmath,amsfonts,amssymb}
\usepackage{float}
\usepackage{mathtools}
\usepackage{colortbl}
\usepackage{tabularx}
\usepackage{ amssymb }

\DeclarePairedDelimiter\floor{\lfloor}{\rfloor}

\title{\LARGE \bf
Dataiku's Solution to SPHERE's Activity Recognition Challenge
}

\author{Maxime Voisin$^{1}$, Leo Dreyfus-Schmidt$^{2}$, Pierre Gutierrez$^{2}$, Samuel Ronsin$^{2}$ and Marc Beillevaire$^{2}$
\thanks{$^{1}$maxime.voisin@stanford.edu}
\thanks{$^{2}$\{leo.dreyfus-schmidt, pierre.gutierrez,samuel.ronsin, marc.beillevaire\} 
\newline \indent @dataiku.com}
\thanks{$^{3}$Second Prize, but third place! The submission of a team ranked top 2 
\newline \indent was deemed invalid}
}

\begin{document}

\maketitle
\thispagestyle{empty}
\pagestyle{empty}


\begin{abstract}
Our team won the second prize$^{3}$ of the Safe Aging with SPHERE Challenge organized by SPHERE, in conjunction with ECML-PKDD and Driven Data. The goal of the competition was to recognize activities performed by humans, using sensor data. This paper presents our solution. It is based on a rich pre-processing  and state of the art machine learning methods. 
From the raw train data, we generate a synthetic train set with the same statistical characteristics as the test set. We then perform feature engineering.
The machine learning modeling part is based on stacking weak learners through a grid searched XGBoost algorithm. Finally, we use post-processing to smooth our predictions over time.\\

\end{abstract}


\section*{INTRODUCTION}

SPHERE organized an activity recognition competition in conjunction with ECML-PKDD and DrivenData. The goal was to recognize activities - postures and movements - from sensor data collected from participants. Our solution reached second prize.

Hopefully, this paper will contain sufficient information to reproduce our results, and we will try to make it transparent when our choices were time-driven rather than following proper scientific method. 

The paper is organized as follow : We first introduce the challenge's goal and its data. 
Then, we describe our approach to create a train set suitable for the machine learning paradigm. The next sections deal respectively with the feature engineering, the machine learning models and the post-processing we used in the competition. 

\section*{CHALLENGE DESCRIPTION}

\subsection*{The target}

The goal of this challenge was to predict, on a second-by-second basis,
a person's activity based on sensor data. It was modeled as a multi-class classification problem.

The target variable could take 20 different values, representing the individual's activities, postures and transitions:  ascend stairs, descend stairs, jump, walk with load, walk, bending, kneeling, lying, sitting, squatting, standing, stand-to-bend, kneel-to-stand, lie-to-sit, sit-to-lie, sit-to-stand, stand-to-kneel, stand-to-sit, bend-to-stand and turn.

Several annotators have manually defined the ground truth for the target variable. For instance, if “jump” is given a value of 0.05 at a given second, this should be interpreted as meaning that on average the annotators marked 5\% of that second as arising from jumping.

\subsection*{Datasets}

For this contest, the SPHERE team equipped a house with three sensor modalities.

\subsubsection*{Accelerometers}
Participants wore a tri-axial accelerometer on their dominant wrist. The device wirelessly transmits the value of acceleration to several receivers positioned within the house. This device gives two valuable pieces of information. First, the value of acceleration, in three directions. Second, the signal power that was recorded by each receiver (in units of dBm) - this data will be informative for indoor localization.

\subsubsection*{Cameras}
Three cameras were used in the living room, hallway and kitchen. Automatic detection of humans was performed. In order to preserve the anonymity of the participants, the raw video data are not shared. Instead, the coordinates of the 2D bounding box, 2D centre of mass, 3D bounding box and 3D centre of mass are provided.

\subsubsection*{Environmental Sensors}
The values of passive (PIR) sensors positioned within the house are given. \\

In order to generate the train data, 10 participants successively performed a script of daily-life actions in this house. Hence, the train data consists of 10 continuous sequences of  monitoring. Each sequence was recorded on a second-by-second basis and lasts approximately 30 minutes. \\

The test data was generated by 10 other participants who followed the same script of daily-life actions: these 10 test sequences of monitoring were also recorded on a second-by-second basis and have a similar duration.

However, instead of supplying 10 continuous test sequences of 30 minutes of monitoring, the SPHERE team randomly split these 10 long sequences into 800 smaller subsequences. To do so, they iteratively sampled a subsequence duration and a number of seconds to drop between two subsequences. The subsequence duration was chosen to follow a uniform distribution between 10 and 30 seconds. The gap length follows a similar distribution.

These subsequences were finally permuted so that it would be difficult to reconstitute the whole 30-minute test sequences. This was probably done  to force the inference of test sequences to be independent of the daily-life action script. The competition entrants' model would thus have to generalize to other scripts and participants, which would make them useful in real-life situations.

\subsection*{Evaluation metric}

Submissions to the competition are evaluated with the Brier score defined as:

$$BS = \frac{1}{N}	\sum_{n=1}^N \sum_{c=1}^C w_c(p_{n,c}-y_{n,c})^2$$

\noindent where $N$ is the number of test sequences, $C$ is the number of classes, $p_{n,c}$ is the predicted probability of instance $n$ being from class $c$, $y_{n,c}$ is the proportion of annotators that labeled instance $n$ as arising from class $c$, and $w_c$ is the weight for each class.

Lower Brier scores indicate better performance, and optimal performance is achieved with a Brier score of 0. Class weights place more weight on the classes that are less frequent.

\section*{PRE-PROCESSING}

\subsection*{Changing the structure of the train and test sets}

The first step was to change the structure of the train set to have its distribution follow that of the test set more closely.
Therefore, we randomly split the 10 train sequences of 30 minutes into 800 smaller subsequences of 10 to 30 seconds, to follow the test set creation methodology. 

By doing this random splitting several times, with different random seeds, it is possible to generate several train sets. Then, we could follow a bagging approach: create one model per train set and average their predictions. This approach showed good results in cross-validation, but due to time constraints it was not part of our final model. \\

The second operation was to optimize a hard target.
In order to have an easier integration with existing Python machine learning libraries (such as scikit-learn and XGboost), we converted our probabilistic (soft) target into a hard target, by keeping for each line the target label with the highest probability.

\subsection*{Cross-validation strategy}

A good cross-validation strategy is crucial to have a faithful estimation of the performance of our models on the leaderboard and to avoid overfitting.
In the competition, train and test sets were generated using two distinct groups of participants. Since it was crucial for our model to generalize to unknown people, we split up the train data into:

\begin{itemize}
\item a train subset: data generated by all individuals but number 6 and 10
\item a validation subset: data generated by individuals 6 and 10\\
\end{itemize}

We observed for each model a rather constant gap between our evaluation and the score obtained on the public leaderboard. So every improvement in our local cross-validation score led to a similar improvement on public leaderboard. 

Note that this cross-validation strategy might not be optimal. It might even cause overfitting if individuals 6 and 10 turn out to be more similar to the individuals involved in public leaderboard, than to those involved in the private one. 



\section*{Feature Engineering}
\subsection*{Initial features}

The raw train and test datasets contained 119 features. However, many of these features are highly correlated or are at a level of granularity too refined. 

For instance, each camera gives the x,y and z-coordinates of the individual's centre of mass. But the camera records at 25 frames per second. 
For every second, we kept the mean, median, min, max and standard deviation of these 25 coordinate values. Hence we got 5 features that describe, for every second, the coordinate values of the individual's centre of mass. Similarly, the accelerometers sample at 20 Hz. So, for a given second of monitoring, we keep the mean, median, min, max and standard deviation of the 5 acceleration values generated by the accelerometer.

\subsection*{Basic feature engineering}

First, we extracted basic features: speeds, accelerations, derivatives of acceleration, second derivatives of acceleration and rotations.

\begin{figure}[H]
\centering
\includegraphics[width=0.5\textwidth]{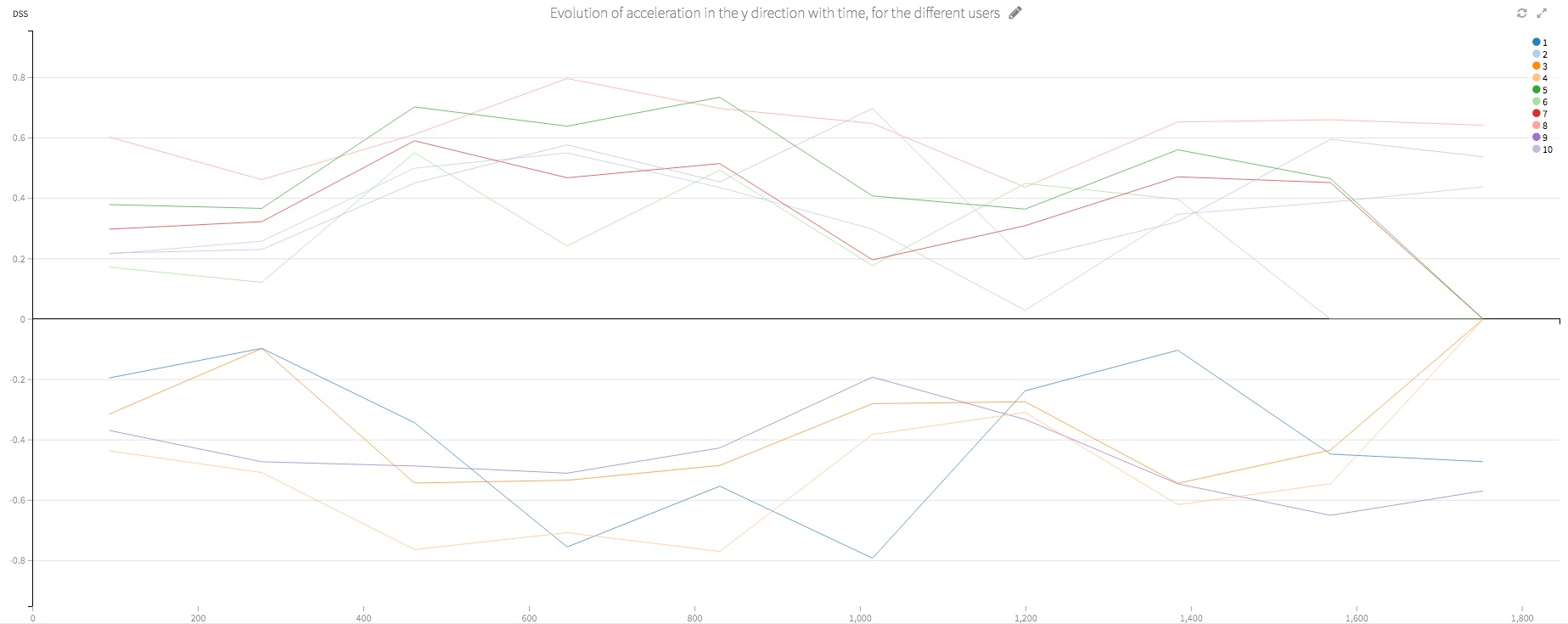}
\caption{The evolution of y-acceleration over time suggests that two distributions coexist in acceleration data}\label{visina8}
\label{fig:acc}
\end{figure}

We noted that the accelerometer was fixed on the individual's dominant wrist. Figure \ref{fig:acc} clearly highlights two distributions of y-accelerations: one for right-handed individuals and another one for left-handed ones. To correct this bias, we multiplied accelerometer x and y data by -1 for left-handed individuals.  \\

\subsection*{Lags and leads}

The previous features exploit the current value of sensors data. But they do not exploit the past or future sensors data. In order to take into account the time series component of the problem, we added lagged variables: values of existing features 1 to 10 seconds before. 
Note how important it is to make train and test sets look alike. In the test set, lagged values are always empty for the first line of each subsequence. Whereas in the train set, if we hadn't split the 30-minute sequences into subsequences, lagged values would hardly be empty. 
We thus avoided a covariate shift on lagged variables. 
\\Experimentally, adding lags helped our model perform well, so we also added leads. 
That is we added new variables giving values of features 1 to 10 seconds after. One may wonder whether adding leads makes sense in real-life applications. Should we really wait a couple of seconds before sending help to an individual, in order to add leads to our model and make sure that the individual actually has a problem?  Yet, given the context of the challenge, we decided to exploit this test set artifact. \\

When looking at Random Forest or Gradient Boosting trees feature importance, we noted that lead variables were as important as lagged ones. Moreover and quite naturally, the importance of the 1 second lag/lead variables was greater than the importance of 2 second ones, etc.

\subsection*{Enriching the data through stack transferring}

The room variable indicates the room where the individual is located. Intuitively, this variable should be very useful to predict activity: for instance, when someone is in the toilets, he is probably not jumping nor lying down.
Unfortunately, this room variable is available in the train set, but missing in the test set. We here propose a technique, that we call \textit{stack transferring}, to propagate this information. It consists in three steps.

\subsubsection*{Step 1 - On the train set, update the room variable: replace its exact values by its out-of-fold predictions}
On the train set, we replace the exact values of the room variable by out-of-folds predictions of the room variable.
In other words, use 9 folds of the train set - corresponding to 9 participants out of 10 - to predict the room variable on the remaining fold. By doing so 10 times, we can predict the room variable on all the train set. This generates out-of-folds predictions of the room variable on the train set.
We can now update the room variable on the train set by dropping the exact values of the room variable and keeping its out-of-fold predictions. 

\subsubsection*{Step 2 - On the test set, predict the room variable}
The room variable being missing on the test set, we can add it as follows. In step 1, we have trained 10 models that predict room variable. We simply apply them to the test set and average these 10 predictions. Now, the room variable should be available on the test set.



\subsubsection*{Step 3 - Use the out-of-folds predictions of the room variable to predict the activity variable}
Now that we have updated the room variable on the train set, and that we have predictions of the room variable on the test set, we can add this variable to the model. It should improve the activity prediction.


Notice that the individuals from train and test sets were asked to perform the same list of actions in the same order. Therefore, the room variable had the same distribution on train and test sets. This is a necessary condition for stack transferring to perform well. \\


Eventually, feature engineering increased the number of variables from 119 to 2700. Table \ref{tab:feat} shows the top 15 features by importance for a Random Forest trained on the engineered train set. 5 of them come from our feature engineering.

\begin{table}[H]
\caption{Top 15 most important features for our level-one random forest learners. Features colored in green come from feature engineering.}
\begin{flushleft}
\begin{tabular}{|c||c|}
\hline
\rowcolor[gray]{0.85} Feature name & Importance (\%)\\
\hline
median acceleration in x direction & 0.5035\\
\hline
std deviation of acceleration in  x direction & 0.4622\\
\hline
max acceleration in  x direction & 0.4290\\
\hline
mean acceleration in x direction & 0.4247\\
\hline
\rowcolor{green} room variable is equal to living room & 0.4119\\
\hline
\rowcolor{green} lead 1 second of mean acceleration in x direction & 0.3726\\
\hline
min y coordinate of 3D centre of mass & 0.3676\\
\hline
\rowcolor{green} lead 1 second of median acceleration in x direction & 0.3463\\
\hline
min acceleration in x direction & 0.3424\\
\hline
min x coordinate of 2D centre of mass & 0.3262\\
\hline
std deviation of acceleration in x direction & 0.3258\\
\hline
max PIR value of the receiver located upstairs & 0.3251\\
\hline
\rowcolor{green} lag 1 second of mean acceleration in x direction & 0.3179\\
\hline
mean acceleration in y direction & 0.3093\\
\hline
\rowcolor{green} length in y direction of 3D bounding box & 0.2991\\
\hline
\end{tabular}
\end{flushleft}
\label{tab:feat}
\end{table}

\section*{ACTIVITY-RECOGNITION MODELS}
\subsection*{Individual models}

It is in general a good idea to start with a simple model that does not need much tuning - for instance a Random Forest - while doing feature engineering. They are easy to implement and able to handle large amounts of variables, so they give valuable feedback on the quality of our work. Feature engineering diminished our Random Forest's error-rate from 22\% to 16.4\%, ranking us $15^{th}$ of the competition.\\

When performance seemed to reach a plateau even when we were adding new features, we tried other models that require more tuning. We then went on for the machine learning blockbuster, XGBoost.
We grid-searched its parameters - max depth, min child weight, column sample by tree, subsample - and derived the optimal number of estimators thanks to an early stopping on users 6 and 10. 
Optimizing XGBoost typically took one hour on our 12 cores computer, which was fast enough to explore a great number of feature combinations.\\

The XGBoost classifier can optimize its predictions for a given loss function. This loss function can be chosen among several pre-implemented loss functions. But the metric of the challenge - Brier score - is not one of them.  So, we chose a random pre-implemented loss function - logloss. It is not an optimal solution, because minimizing logloss should not necessarily lead to minimizing the Brier score. However, this already performed very well: our error rate reached 14.6\% and ranked us top 5.

We could then try to customize the XGBoost code to make it optimize the Brier score loss function instead of logloss.

\subsection*{Customizing XGBoost}
Our goal was to make the XGBoost classifier optimize its predictions for the metric of the challenge - the Brier score. XGBoost provides a Python API to  customize softmax loss functions, by defining their gradient and hessian. The first step was to define the softmax Brier score loss function:

\begin{flalign*}
\mathcal{L}(\mathbf{p}) 
=
\frac{1}{N} \sum_{n=1}^N \sum_{c=1}^C w_c(
\sigma_{n,c}(\mathbf{p}) 
- y_{n,c})^2
\end{flalign*}

\noindent where $\mathbf{p}$ and $\sigma_{n,c_0}(\mathbf{p})$ are respectively equal to
$$\mathbf{p}=\Big((p_{n,c})_{\substack{1\leq n\leq N \\ 1\leq c\leq C}}\Big),    \sigma_{n,c_0}(\mathbf{p})=\frac{e^{p_{n,c_0}}}{\sum_{c}e^{p_{n,c}}}$$

\noindent We can then implement the loss gradient and hessian based on the following expressions. Notice that XGBoost does not work with the exact hessian but with its diagonal approximation.
\begin{flalign*}
\frac{\partial\mathcal{L}}{\partial p_{n,c_0}}(\mathbf{p})
= \frac{2}{N} \sum_{n=1}^N \sigma_{n,c_0}(\mathbf{p})
\Big[ \sum_{c=1}^C w_c\big[y_{n,c} - \sigma_{n,c}(\mathbf{p})\big]\sigma_{n,c}(\mathbf{p}) &&\\
- w_{c_0}\big[y_{n,c_0} - \sigma_{n,c_0}(\mathbf{p})\big] \Big]
\end{flalign*}

\begin{flalign*}
\frac{\partial^2  \mathcal{L}}{\partial p_{n,c_0}^2}& ( \mathbf{p})
= \frac{2}{N} \sum_{n=1}^N \Big[  w_{c_0} \big[y_{n,c_0}\sigma_{n,c_0}(\mathbf{p}) -2y_{n,c_0}\sigma_{n,c_0}^2(\mathbf{p}) &\\ 
&+2\sigma_{n,c_0}(\mathbf{p})^2+4\sigma_{n,c_0}^3(\mathbf{p})\big]&\\ 
&+\sum_{c=1}^Cw_c \sigma_{n,c_0}(\mathbf{p}) \sigma_{n,c}(\mathbf{p})\big[y_{n,c}\sigma_{n,c_0}(\mathbf{p})&\\ 
&-3\sigma_{n,c_0}(\mathbf{p})\sigma_{n,c}(\mathbf{p})+\sigma_{n,c}(\mathbf{p})\big] \Big] &
\end{flalign*}

Unfortunately, the XGBoost Python API only allows this easy customization of loss function when the target is a hard target. In the SPHERE Challenge, the target is probabilistic. An easy way to deal with this issue was to convert the probabilistic (soft) target into a hard target, by keeping for every line the label that has the highest probability. This inevitably generated an approximation in the metric optimized by XGBoost. We managed to minimize this approximation by duplicating lines on our train dataset. For instance, for a given line, if label A has a probability of 0.7, label B of 0.1 and label C of 0.2, then we would create K new lines: $\floor*{0.7K}$ lines that would have label A as hard target, $\floor*{0.1K}$ line would have label B as hard target and $\floor*{0.2K}$ lines would have label C as hard target. By doing so, our XGBoost would optimize the following approximate softmax Brier score:

\begin{flalign*}
\mathcal{L}_{approx}(\mathbf{p}) = \frac{1}{N}	\sum_{n=1}^N \sum_{c=1}^C w_c(y_{n,c}^2- 2y_{n,c}\frac{\floor*{K\sigma_{n,c}(\mathbf{p}) }}{K} &&\\
+
\frac{\floor*{K\sigma_{n,c}(\mathbf{p})}^2}{K^2} )
\end{flalign*}

which is quite close to the exact softmax Brier score:
$$ \mathcal{L}_{exact}(\mathbf{p}) = \frac{1}{N} \sum_{n=1}^N \sum_{c=1}^C w_c(y_{n,c}^2 -2y_{n,c}\sigma_{n,c}(\mathbf{p}) +\sigma_{n,c}(\mathbf{p}) ^2)$$

We refer to K as a resolution parameter, that governs the approximation in the Brier score metric made by XGBoost when dealing with a hard target instead of a probabilistic one. Higher values of K reduce this approximation. We have implemented this method with $K=10$, meaning that our train dataset consisted of 100,000 lines and 2700 features: but XGBoost training time was too long.	
\\

Therefore, the only solution was to fork the C++ XGBoost source code to make it accept customized loss functions even when the target is probabilistic. This was much trickier, but unsurprisingly it gave slightly better results than the traditional XGBoost. Customized XGBoost decreased our CV score from 0.1817 to 0.1814.

\subsection*{Stacking}
Once we had trained 10 individual models - including linear regressions, Naive Bayes classifiers, Random Forests, extra-trees and XGBoost models -, we opted for ensemble learning methods. A grid-searched XGBoost combined the predictions of our individual models and leveraged their strengths. It turned out to be very efficient: it reduced our error rate to 12.9\% and ranked us number 1 at that point.

\begin{figure}[H]
\centering
\includegraphics[width=0.5\textwidth]{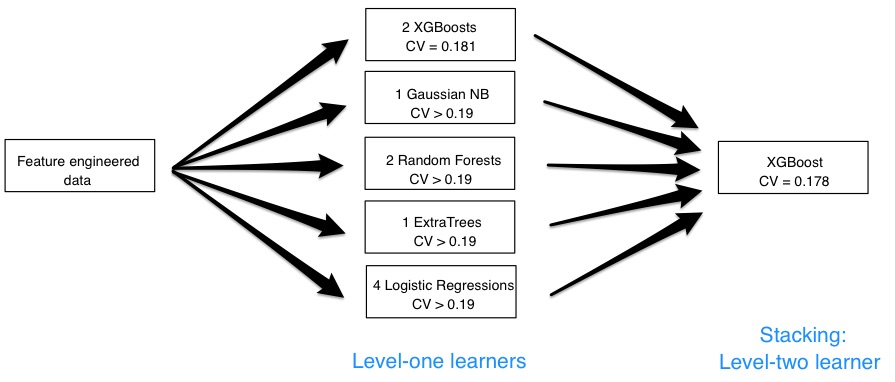}
\caption{The stacking technique improved our CV results from 0.181 to 0.178 }\label{visina8}
\end{figure}
\section*{POSTPROCESSING}

The previous approaches consider each prediction independently. However, it seems very unlikely that a person lying on a bed can be jumping the next second. This means that there are chances that transitions from one activity to another follow different probabilities. This mathematical property is known as the Markov chain property. A great way to take advantage of this underlying structure is to implement Hidden Markov Models. 

\subsection*{Smoothing predictions over time}

Yet, given the deadline, we did not have time to implement HMM models. We rather opted for a post-processing that smooths predictions over time. The idea is to make a weighted average between the activity predictions of a given second and the activity predictions of the last two seconds and of the future two seconds. 
We optimized the coefficients of this weighted average. Post-processing gave tremendous cross-validation results, with an error rate around 11\%. However, we did not have time to submit it in our final model.

\section*{CONCLUSION}

In this paper, we presented our solution to SPHERE Challenge as well as several techniques that may have worked if we had more time. Our final solution is based on a rich pre-processing and cutting-edge machine learning methods.

After recreating a train set similar to the test set, we perform feature engineering. To our knowledge, what we call "stack transferring" - the idea of using predictions of a variable known in the train set but not in the test set as features - is new. 

The final model is based on the stacking of weak learners through a grid searched XGBoost algorithm.  \\

Our solution won the second prize of the challenge on the private leader-board, though we were ranked first on the public one. We hope that this work can modestly contribute to finding better way to detect old people fall for a quicker intervention. 

\addtolength{\textheight}{-12cm}   



\section*{ACKNOWLEDGMENT}
 
We thank Dataiku for allocating time and servers for our team. We also thank DrivenData, ECML-PKDD and SPHERE for organizing and hosting this contest and supplying these valuable datasets. \\


\end{document}